\documentclass[journal]{IEEEtran}

\usepackage{epsfig}
\usepackage{graphicx}
\usepackage{amsmath}
\usepackage{amssymb}
\usepackage{enumitem}
\usepackage{bm}
\usepackage{booktabs}
\usepackage{subcaption}
\usepackage{multirow}

\newcommand{\etal}{\textit{et al}. }
\newcommand{\ie}{\textit{i}.\textit{e}.}
\newcommand{\eg}{\textit{e}.\textit{g}.}

\makeatletter
\newcommand*\bigcdot{\mathpalette\bigcdot@{.7}}
\newcommand*\bigcdot@[2]{\mathbin{\vcenter{\hbox{\scalebox{#2}{$\m@th#1\bullet$}}}}}
\makeatother

\usepackage{mathtools}

\usepackage{tabularx} 
\usepackage{adjustbox}

\ifCLASSOPTIONcompsoc
  \usepackage[nocompress]{cite}
\else
  \usepackage{cite}
\fi

\begin{document}
%
\title{Adaptive Affinity for Associations in \\Multi-Target Multi-Camera Tracking}
%
%
%

\author{Yunzhong Hou,
        Zhongdao Wang,
        Shengjin Wang,
        and Liang Zheng
\IEEEcompsocitemizethanks{\IEEEcompsocthanksitem Yunzhong Hou and Liang Zheng are with Research School of Computer Science, Australian National University, Canberra, ACT 2601, Australia. \protect\\
E-mail: \{firstname.lastname\}@anu.edu.au
\IEEEcompsocthanksitem Zhongdao Wang and Shengjin Wang are with State Key Laboratory of Intelligent Technology and Systems, Tsinghua National Laboratory for Information Science and Technology, Department of Electronic Engineering, Tsinghua University, Beijing, China. 
E-mail: wcd17@mails.tsinghua.edu.cn, wgsgj@tsinghua.edu.cn.
}
\thanks{This work was supported by the ARC Discovery Early Career Researcher Award (DE200101283) and the ARC Discovery Project (DP210102801).}
}

%
%

\markboth{Journal of \LaTeX\ Class Files}%
{Hou \MakeLowercase{\textit{et al.}}: Adaptive Affinity for Associations in Multi-Target Multi-Camera Tracking}
%



\maketitle

\begin{abstract}
Data associations in multi-target multi-camera tracking (MTMCT) usually estimate affinity directly from re-identification (re-ID) feature distances. However, we argue that it might not be the best choice given the difference in matching scopes between re-ID and MTMCT problems. Re-ID systems focus on \textit{global matching}, which retrieves targets from all cameras and all times. In contrast, data association in tracking is a \textit{local matching} problem, since its candidates only come from neighboring locations and time frames. 
In this paper, we design experiments to verify such misfit between \textit{global} re-ID feature distances and \textit{local} matching in tracking, and propose a simple yet effective approach to adapt affinity estimations to corresponding matching scopes in MTMCT. Instead of trying to deal with all appearance changes, we tailor the affinity metric to specialize in ones that might emerge during data associations. To this end, we introduce a new data sampling scheme with temporal windows originally used for data associations in tracking. Minimizing the mismatch, the adaptive affinity module brings significant improvements over global re-ID distance, and produces competitive performance on CityFlow and DukeMTMC datasets. 

\end{abstract}


%
\IEEEpeerreviewmaketitle

\section{Introduction}
\IEEEPARstart{M}{ulti}-target multi-camera tracking (MTMCT) aims to 
formulate trajectories for different identities across multiple cameras. 
It plays a vital role in many applications including smart city analysis and autonomous driving~\cite{ristani2016performance,tang2019cityflow}.

MTMCT extends the multiple object tracking (MOT) problem, which only focuses on a single camera and does not need to align cross camera identity. In fact, MTMCT can be divided into two steps: first, in single camera tracking (SCT), trajectories are linked within each camera; second, in multiple camera tracking (MCT), within-camera trajectories are associated across cameras. Most existing works on MOT and MTMCT follow the tracking-by-detection paradigm~\cite{andriluka2008people}, where data association is arguably the most defining part (object detection is separately studied in other works~\cite{girshick2015fast,ren2015faster,liu2016ssd}). 
In data association, targets of the same identity are linked into trajectories based on cost matrices using graph optimization. 
Such cost matrices are generated by affinity or similarity estimation, and their quality greatly influences the data association performance~\cite{fjallstrom1998algorithms,milan2016mot16}. For affinity estimation, re-identification (re-ID) \cite{zheng2015scalable} feature distance is widely chosen in tracking systems \cite{yu2016poi,tang2017multiple,ristani2018features,tang2019cityflow}, since it can distinguish target identities across multiple cameras. 

\begin{figure*}
\centering
    \begin{subfigure}[b]{\linewidth}
    \centering
        \includegraphics[width=\textwidth]{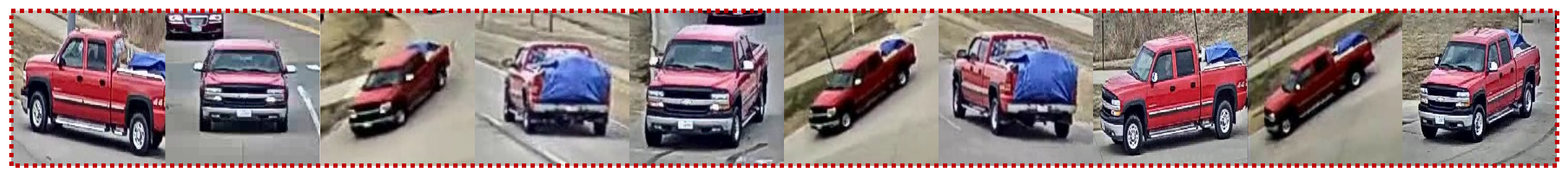}
        \caption{Re-identification (re-ID)}
    \end{subfigure}
    \begin{subfigure}[b]{\linewidth}
    \centering
        \includegraphics[width=\textwidth]{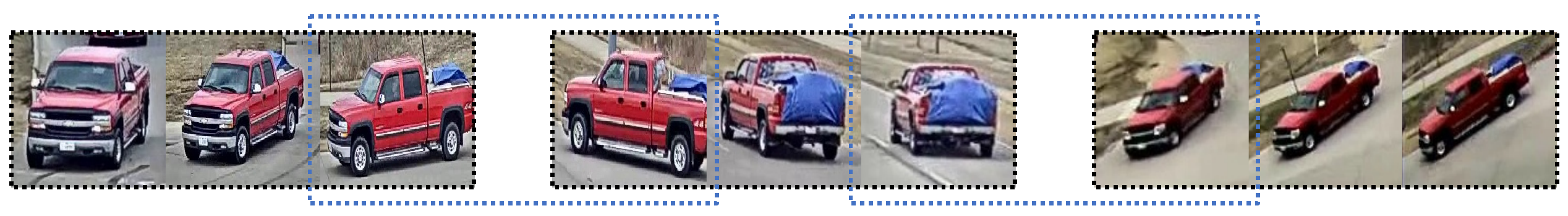}
        \caption{Multi-target multi-camera tracking (MTMCT)}
    \end{subfigure}
\caption{Differences in matching scopes between re-ID and MTMCT. Re-ID (red dotted box) aims to retrieve from all locations and all times, thus requiring re-ID features to deal with all potential appearance changes at the same time. 
In contrast, MTMCT usually associates candidates in an hierarchical manner, by first tracking within single camera (black dotted box) and then across multiple cameras (blue dotted box). With limited problem sizes (matching scopes), associations in tracking only have to deal with a limited number of appearance changes at a time. 
Given the difference between matching scopes, we argue that directly using re-ID feature distance as affinity might not be the best choice for MTMCT data association. 
}
\label{fig:intro}
\end{figure*}

However, in this work, we find that \textit{directly} using the re-ID feature distance as affinity may not be the best choice for MTMCT. 
As shown in Fig.~\ref{fig:intro}, re-ID is a \textit{global matching} problem, where the system tries to retrieve targets of the same identity from anywhere any time. Thus, re-ID systems have to deal with \textit{all} possible appearance changes (\eg, occlusions, pose changes) \textit{at the same time}. 
On the other hand, tracking aims to formulate continuous trajectories, where data associations are usually conducted in an hierarchical (first SCT and then MCT) and iterative (iteratively adding new candidates to existing trajectories) manner with smaller matching scopes (one frame or one time-batch at a time). As such, associations in tracking can be regarded as \textit{local matching} problems, and the appearance changes that the system has to deal with are \textit{limited} to the problem sizes considered in associations (matching scopes). 
The difference in problem sizes (\textit{global matching} for re-ID versus \textit{local matching} for tracking) violates the common belief that affinity estimations should be tailored for the corresponding matching scopes. 

This phenomenon (different problem sizes in re-ID features and MTMCT data associations) is previously \textit{overlooked} in MTMCT problems, and is \textit{less pronounced} in MOT problems. In fact, in MOT (not to be confused with SCT in MTMCT), targets for re-ID feature learning only come from a single camera within a limited time period, and contain a similar level of appearance changes as the tracking system might encounter during data associations. 
By contrast, in MTMCT, such re-ID training data might come from different cameras over a long period of time (for them to travel to different camera locations), and contains much larger appearance changes. During data association, such drastic appearance changes are unlikely to all appear at the same time in either SCT or MCT, since the association problem sizes are usually limited to neighboring time frames and cameras.

In this paper, we first design experiments to verify this mismatch between \textit{global} re-ID features for affinity estimation and \textit{local} matching in MTMCT data association. To this end, we skip the detection and graph optimization algorithms, and directly compare the estimated affinity (similarity/dissimilarity) between the ground truth bounding boxes. Experiments show that directly using the same re-ID feature distance for both SCT and MCT can result in higher ratios of false positives, suggesting that the \textit{global affinity} scores might be too tolerant and not discriminative enough for the \textit{local matching}. 

To address this mismatch, we propose a simple-yet-effective adaptive affinity module for data associations in MTMCT that tailors affinity estimations to the corresponding matching scopes. 
In order to benefit from the strong identification ability of re-ID features while tailoring them to fit the local matching problem, we learn metric networks on top of the re-ID features. 
To fully exploit existing Siamese networks~\cite{bromley1993signature}, we propose new data sampling mechanisms for metric leaning.
To produce affinity estimations that best fit the smaller matching scopes in tracking, we train the metric networks with data pairs that might appear in the corresponding association problems. To this end, during training, we select data pairs with \textbf{the same} temporal windows adopted in the data association step. 
Specifically, we learn an intra-camera metric for associations in SCT and an inter-camera metric for associations in MCT. 
For intra-camera metric, temporal windows select data pairs within the same camera; and for inter-camera metric,  data pairs from all cameras are allowed, and temporal windows automatically select those in neighboring cameras. 
We clarify that this work does not make any architecture-wise contribution or design any new trackers. Its key contributions are 1) \textbf{identifying and exploring} the mismatch between global re-ID feature distances and local matching in MTMCT data associations and 2) a new \textbf{data sampling scheme} for learning adaptive affinity metrics to bridge the mismatch.

We show that the proposed adaptive affinity can effectively improve tracking accuracy on two MTMCT datasets, including a vehicle dataset, CityFlow~\cite{tang2019cityflow}, and a pedestrian dataset, DukeMTMC~\cite{ristani2016performance}. It can also be applied and on top of multiple re-ID features, such as IDE \cite{zheng2016person}, PCB \cite{sun2018PCB} and the triplet feature \cite{hermans2017defense}. 
With a competitive tracker \cite{ristani2018features}, we report the state-of-the-art accuracy on the DukeMTMC dataset.

\section{Related Work}\label{related work}

\textbf{Multi-object tracking.}
Multi-object tracking (MOT) ~\cite{leal2015motchallenge,milan2016mot16,leal2017tracking} tracks multiple targets within each scenario. 
The MOT challenge and its datasets witnessed the bloom of the modern MOT system~\cite{leal2015motchallenge,milan2016mot16}. 
Most MOT systems follow the tracking-by-detection paradigm~\cite{andriluka2008people}. Since the detection part is also studied in other field~\cite{ren2015faster,liu2016ssd}, many researchers focus on data association methods for MOT systems. 
For \textit{affinity estimation}, existing works adopt convolutional neural network (CNN) feature distance~\cite{kim2015multiple,yang2017hybrid} or Siamese network scores~\cite{zhang2016tracking,son2017multi}. 
For \textit{optimization algorithms}, there are both online and offline ones. Online tracking methods have a very small association problem size as they only consider the current frame and thus have minimal computation cost~\cite{fagot2016improving,choi2015near}. As targets move continuously, their trajectories can still be formulated iteratively even with per-frame matching in online systems. 
The offline methods, on the other hand, consider multiple frames inside a temporal window during association. The increase in problem size can improve tracking performance at the cost of higher computation complexity. 
They usually formulate the problem as batch optimization, such as shortest path \cite{berclaz2011multiple,dehghan2015target}, bipartite graph~\cite{brendel2011multiobject,cai2014exploring}, and pairwise terms~\cite{wang2014tracklet,yu2016solution}. To reduce computation complexity, some employ a hierarchical approach~\cite{singh2008pedestrian,shitrit2014multi}, or temporal sliding windows~\cite{sadeghian2017tracking,choi2015near}. 

\textbf{Multi-target multi-camera tracking.}
Multi-camera monitoring receive wide attention from researchers \cite{d2015survey,tahir2014low,ferecatu2009multi,di2016novel,black2006multi,hou2020multiview,hou2021multiview}. 
Specifically, multi-target multi-camera tracking (MTMCT) tracks targets across cameras \cite{tesfaye2019multi,maksai2017non,ristani2018features,yoon2018multiple,zhang2017multi,jiang2018online}, and assumes no overlapping field-of-view across cameras. Similar to MOT systems, MTMCT systems also follow the tracking-by-detection paradigm. 
For \textit{affinity estimation}, most existing works also directly adopt the CNN feature distance trained from re-ID systems~\cite{ristani2018features,jiang2018online,tang2019cityflow}. With that said, in contrast to MOT problems, targets in MTMCT systems can appear in more than one camera/scenario, leading to a potential misfit between the affinity and the association problem sizes. 
Speaking of \textit{association algorithms}, given the huge problem size, existing works on MTMCT usually solve it in a hierarchical manner. Specifically, first, single camera tracking formulates within camera trajectories. Second, cross camera tracking link the trajectories across cameras. Temporal windows are usually adopted to further restrict the problem size, so as to maintain a manageable problem size. 
In order to solve it in an online manner, Yoon \etal \cite{yoon2018multiple} formulate the problem as track-hypothesis trees and solve it via multiple hypothesis tracking algorithms. 
On the other hand, offline methods \cite{ristani2018features,ristani2016performance,tesfaye2019multi,zhang2017multi} employ batch optimization techniques for higher accuracy, which is similar to MOT trackers. 
For example, in~\cite{maksai2017non}, Maksai \etal propose a global optimization method via a non-Markovian problem formulation. Tesfaye \etal provide a quadratic optimization formulation with constrained dominant sets clustering techniques~\cite{tesfaye2019multi}.
Vehicle MTMCT is also studied. Tang \etal~\cite{tang2018single} use multiple cues to accommodate the similar appearance, heavy occlusion, and large viewing angle variation in vehicle tracking. 

\textbf{Re-identification.}
The affinity or similarity for data association in tracking are also studied \cite{turaga2009maximin,sahbi2018learning,yang2012affinity,wang2015global,zhou2016similarity}. With that said, most recent tracking systems adopts re-identification (re-ID) distance for similarity estimation.
Re-ID systems focus on retrieving all targets of the same identity across cameras. 
CNN based methods achieve very high accuracy in pedestrian re-ID \cite{zheng2015scalable,sun2018PCB,wang2019learning,ye2021deep}. Ye \etal \cite{ye2021deep} provide a detailed investigation to state-of-the-art re-ID methods and their application. Multiple loss functions are proposed towards training better re-ID models, such as the contrastive loss~\cite{varior2016gated} and triplet loss~\cite{schroff2015facenet,cheng2016person,liu2017end}. Hermans \etal investigate training techniques and propose hard negative mining~\cite{hermans2017defense} for triplet loss. Zhong \etal propose random erasing as a data augmentation method to enrich the database \cite{zhong2017random}. Wang \etal investigates deep hidden attributes for further performance increases. 
Vehicle re-ID, on the other hand, also attracts much attention~\cite{wang2017orientation,zhou2018aware,tang2019cityflow}. Compared to the pedestrian counterpart, vehicle re-ID exhibits additional challenges as the targets might look very similar. 
Video re-ID use tracklet feature to represent the video containing the target \cite{zheng2016mars}. Spatial-temporal cues also help re-ID~\cite{liu2015spatio,li2018diversity,liu2019spatial,chen2019spatial}, but using them do not change the global matching nature of re-ID. 

Departing from existing works, this paper studies the intrinsic dissimilarities between MTMCT and re-ID. Instead of network architectures or tracker designs, we propose a new training data sampling method, which adapts global re-ID features to affinity metrics that  suit local matching in MTMCT data associations.

\section{Affinity and Association}

In this section, we first introduce affinity estimations and data associations in MTMCT. Then, we design experiments to verify the mismatch in problem sizes between re-ID feature distances and MTMCT data associations. 

\subsection{Affinity Estimation}
\label{secsec:re-id-affinity}

Similar to many previous works~\cite{ristani2018features,zhang2017multi,jiang2018online,tang2019cityflow}, we first calculate the affinity $a_{i,j}$ from re-ID feature distances,
\begin{align}
\label{eq:w_a}
    a_{i,j} = \frac{\mathit{thres}-dist\left(\bm{f}_i, \bm{f}_j \right)}{\mathit{thres}},
\end{align} 
where $\bm{f}_i$ and $\bm{f}_j$ denote the re-ID feature for target $i$ and target $j$, respectively. $dist\left(\cdot, \cdot\right)$ denotes the distance function, where we choose the Euclidean distance. $thres$ denotes the threshold for assuming the data pair as of the same identity. We denote data pairs with different identity as negative, and data pairs of the same identity as positive. Following Ristani \etal~\cite{ristani2016performance}, we choose $\mathit{thres} = \frac{\mu_n+\mu_p}{2}$, where $\mu_p$ and $\mu_n$ denote the average feature distance of positive and negative data pairs, respectively. We calculate $\mu_p$ and $\mu_n$ from \textit{all} possible data pairs following the global retrieval task of re-ID. In this manner, positive data pairs should have positive affinity scores, and negative data pairs should have negative affinity scores.

\subsection{Data Association}
\label{secsec:association}

Similar to previous works~\cite{ristani2016performance,ristani2018features,tang2019cityflow}, we conduct the data association for MTMCT in a hierarchical and iterative manner. 
Hierarchical means that the detection bounding boxes are first connected into within camera trajectories in the single camera tracking (SCT) step, and then these within camera trajectories are linked across multiple cameras in the multiple camera tracking (MCT) step. 
Iterative means that the data association problem sizes are limited to the size of temporal windows. Such temporal windows include detection bounding boxes within a single camera for SCT, and within camera trajectories for MCT. 

With a limited association problem size, we then conduct the graph optimization problem for the association. Specifically, we create a graph with the targets as nodes, and their affinity estimations (Eq.~\ref{eq:w_a}) as weighted edges. For all targets within a temporal window in either SCT or MCT, we optimize the following problem,
\begin{align}
     &\max_{x_{i,j}}{\sum_{i,j}{x_{i,j}a_{i,j}}}, 
\label{eq:optimize}
\end{align}
where $x_{i,j}\in \left\{-1,1\right\}$ is an indicator for whether target $i$ and target $j$ are of the same identity. Given perfect affinity estimations (positive affinity scores for all data pairs of the same identity and negative scores otherwise), maximizing Eq.~\ref{eq:optimize} should return perfect association results. On the other hand, with affinity estimations of undesired quality, the graph optimization also struggles, clearly showing the importance of a good affinity estimation.

\subsection{Verifying the Mismatch}
\label{secsec:mismatch}

The re-ID feature distance specializes in the global matching problem in re-ID, which deals with all possible appearance changes at the same time. On the other hand, data associations in MTMCT are usually hierarchical and iterative, which limits the matching scopes and reduces the number of possible appearance changes each time. In this paper, we argue that the global re-ID feature distance as affinity is not the best choice for local matching in MTMCT, and design experiments to verify this phenomenon. 

\begin{figure}[t]
\centering
\includegraphics[width=\linewidth]{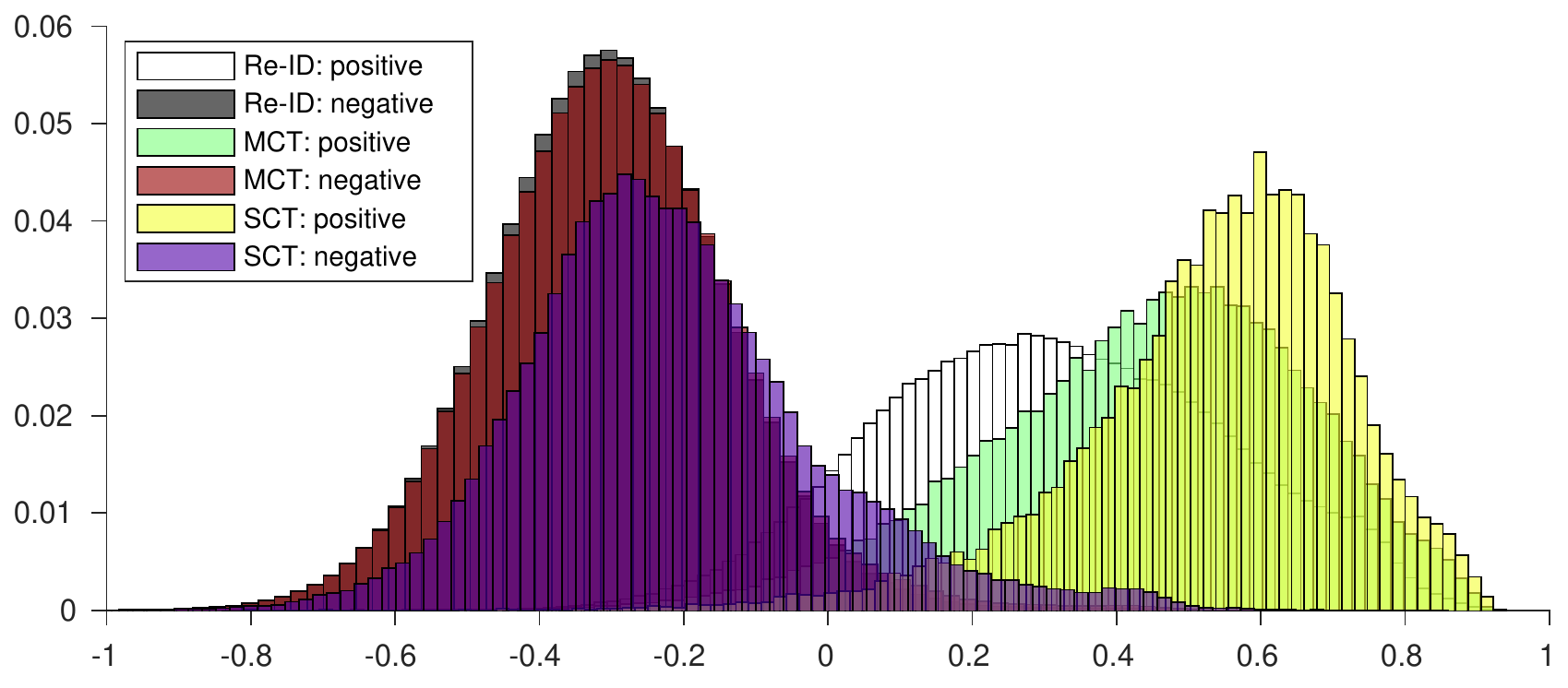}
\caption{
Normalized distributions of the pairwise affinities under different matching scopes on CityFlow test set.
}
\label{fig:affinity_dist}
\end{figure}

\textbf{Preliminary experiment design.} 
To evaluate the affinity estimation, we skip the detection step and the graph optimization step. We consider the ground truth bounding boxes in the validation partition, which are not accessible during the re-ID feature learning. We use this preliminary experiment setting in Fig.~\ref{fig:affinity_dist}, Table~\ref{tab:affinity_dist}, and Fig.~\ref{fig:fpfn}. 
For positive and negative data pairs, we further inspect their matching errors by classifying them into true positives (TP), true negatives (TN), false negatives (FN), and false positives (FP) according to their affinity scores. We also report the overall success (True = TP + TN) and failure (False = FP + FN) rates.

\textbf{Results.} 
We compare the affinity score distribution from different matching scopes in re-ID, MCT, and SCT. From re-ID to MCT to SCT, the matching scopes become smaller and smaller (re-ID considers targets from all cameras and all times; MCT considers targets across multiple cameras within a longer temporal window; SCT only considers targets within the same camera inside a shorter temporal window). 

In Fig.~\ref{fig:affinity_dist}, we show normalized distributions of positive pairs and negative pairs. Going from re-ID to MCT to SCT, as the problem sizes decrease, the distributions of affinity scores in Eq.~\ref{eq:w_a} move towards the right-hand side, indicating higher affinity values on average and more false positives. The global re-ID distance based affinity score (Eq.~\ref{eq:w_a}) do a great job in separating the positive and negative pairs in re-ID problems, but are not as effective in the MCT and SCT problems (cannot effectively distinguish the positive and negative pairs in those scenarios). For example, for SCT, within each camera, re-ID features of different identities are still relatively similar~\cite{zhang2020single}, leading to false positives. For MCT, since the temporal sliding windows only include targets from neighboring cameras, the re-ID features also have limited diversities, and can lead to high false positive rates. 

This is further proven by the statistics of the positive and negative pairs in different matching problems in Table~\ref{tab:affinity_dist}. Going from re-ID to MCT to SCT, as the matching scopes decrease, using the same global re-ID distance based affinity score (Eq.~\ref{eq:w_a}) leads to a higher failure rate in differentiating the positive and negative pairs.
Specifically, we witness stable false negatives and a lot more false positives in data associations for MTMCT (SCT and MCT). 

\begin{table}[t]
\caption{
Percentages (\%) of positive (P) and negative (N) data pairs on CityFlow test set. We also show matching errors with true positives (TP), true negatives (TN), false negatives (FN), false positives (FP), and overall success (True) and failure (False) rates.
}
\label{tab:affinity_dist}
\centering
\begin{tabular}{l|cc|cccc|cc}
\toprule
      & P    & N    & TP   & TN   & FN  & FP   & True & False \\ \hline
re-ID & 0.8  & 99.2 & 0.7  & 96.5 & 0.1 & 2.8  & 96.6 & 2.9   \\ \hline
MCT   & 2.9  & 97.1 & 2.8  & 93.3 & 0.1 & 3.8  & 93.4 & 3.9   \\ \hline
SCT   & 15.7 & 84.3 & 15.4 & 73.3 & 0.2 & 11.1 & 73.5 & 11.3  \\ \bottomrule
\end{tabular}
\end{table}

From Fig.~\ref{fig:affinity_dist} and Table~\ref{tab:affinity_dist}, we verify that directly using re-ID feature distance as affinity might not be the best choice for data associations in MTMCT. We find the global re-ID distance failed to do as good a job in differentiating the positive and negative pairs in SCT and MCT as in re-ID. This is not a problem of re-ID features, since the re-ID problem is essentially different from MTMCT in terms of matching scopes, which requires the feature to be robust and can deal with all appearance changes at once. Even if we adopt re-ID features with higher performance, their higher overall robustness does not necessarily translate into being more discriminative in data associations in MTMCT (see Section~\ref{secsec:variants} and Table~\ref{tab:duke_val} for more details). Instead, what we need is affinity estimations more suitable for the data association problems in MTMCT.

\textbf{The mismatch and its non-significance in MOT problems.}
It is noteworthy that the problem scale mismatch between re-ID features and MTMCT data associations is less pronounced in multiple object tracking (MOT) problems, where the target only appears in a single camera. Re-ID features learned on MOT data only learns to deal with limited appearance changes within a single camera and a short time period, which are of similar levels to that in the MOT data association problem. In this case, the problem scales between re-ID feature learning and MOT data association are already very similar, and there does not exist a significant mismatch between the matching scopes. 


\begin{figure*}[t]
    \centering
    \includegraphics[width=\linewidth]{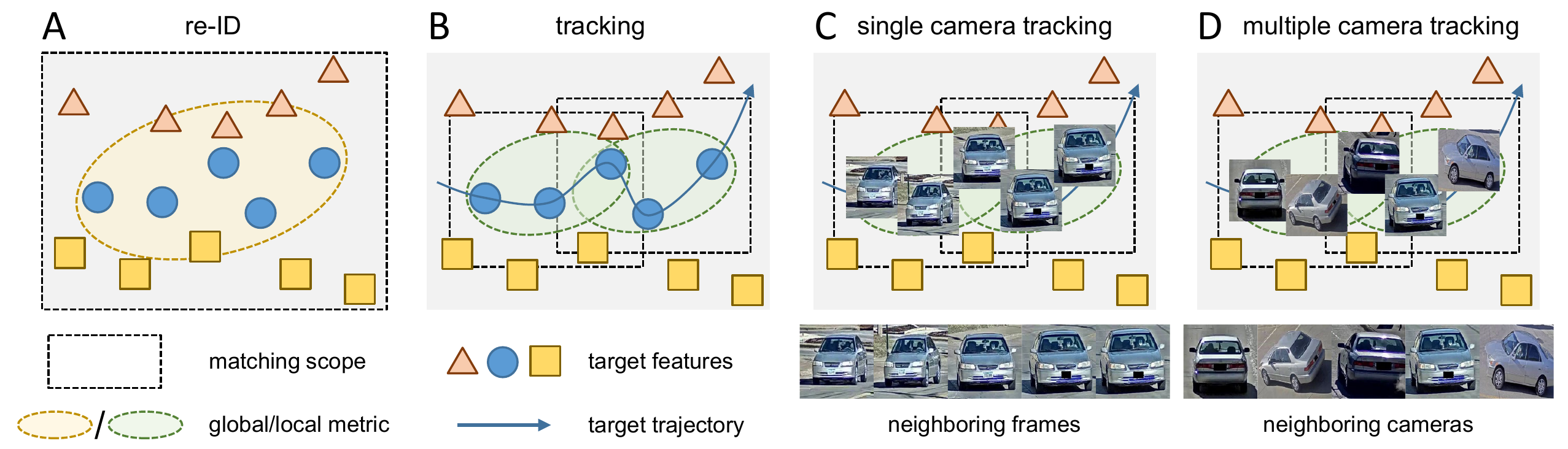}
    \caption{
    \textbf{(A)} Re-ID system matches globally, and global re-ID feature distances as affinity focus on all data pairs (Eq.~\ref{eq:w_a}).  
    \textbf{(B)} Matching scopes in tracking are usually smaller (Section~\ref{secsec:association}). Global metric from re-ID might have limited performance due to its slack decision boundary, and potentially introduces more false positives (Section~\ref{secsec:mismatch}). Adaptive affinities can learn tighter decision boundaries, and should be a better fit for local matching in tracking. 
    The proposed adaptive affinity module learns \textbf{(C)} an intra-camera metric for SCT and \textbf{(D)} an inter-camera metric for MCT. Similar to their corresponding matching scopes, the former is learned from neighboring frames in the same camera, and the latter is learned on tracklet pairs across neighboring cameras. 
    In this manner, we make sure that the affinity estimations are tailored for the association matching scopes. 
    }
    \label{fig:method}
\end{figure*}

\section{Method}

In this section, in order to tailor affinity estimations for corresponding data association problem sizes, we present a simple-yet-effective adaptive affinity module for data associations in MTMCT. 
The \textit{key contribution} lies in the proposed data sampling schemes for learning such adaptive affinity. 
To highlight the benefit of our data sampling scheme, we adopt existing metric learning techniques, which also makes the overall method easy-to-apply. 

\subsection{Intra-Camera Metric and Inter-Camera Metric}
\label{secsec:two_metrics}

Learned from all cameras, the re-ID features possess strong identification ability. However, as mentioned in Section~\ref{secsec:mismatch}, directly adopting global re-ID feature distance might not be the best choice for local matching problems in MTMCT. To benefit from the strong ability of re-ID features while tailoring them to fit the smaller matching scopes in tracking, we learn metric networks on top of re-ID features for affinity estimation. 

Specifically, we learn an intra-camera metric for associations in SCT and an inter-camera metric for associations in MCT, respectively (Fig. \ref{fig:method}). 
Different form re-ID feature distance that focuses on large matching scopes (all data pairs), the proposed adaptive affinity focuses on smaller matching scopes in MCT or SCT (data pairs within temporal windows). 
To adapt re-ID features to smaller matching scopes, we train the metric network with data samples that might appear in corresponding association problems in tracking. 
To this end, we sample data pairs with temporal windows from SCT and MCT data associations and train intra-camera metric and inter-camera metric for SCT and MCT, respectively. 

During training, we select positive/negative pairs with a $1:1$ ratio for data balance, and feed the absolute difference vector $\bm{f}=\left|\bm{f}_i - \bm{f}_j\right|$ into the metric network as input.

\textbf{Intra-camera metric.} 
For data associations in SCT, we train an intra-camera metric to provide similarity estimation between data pairs. 
In training, we sample data pairs within a small temporal duration of $\tau_\text{S}$ within each camera.

\textbf{Inter-camera metric.}
For data association in MCT, we train an inter-camera metric to provide similarity estimation between single camera trajectories. 
Specifically, for positive data pairs, within the $\tau_\text{M}$-sized window, we choose from targets of the same identity but from different cameras; 
for negative data pairs, within the temporal sampling window $\tau_\text{M}$, we choose randomly from all cameras. 

\textbf{Sampling window lengths} are critical hyper-parameters in the proposed adaptive affinity module, and are set differently 1) for different datasets and 2) for SCT and MCT in the same dataset. To achieve the best fit, during training, we set single camera sampling window length $\tau_\text{S}$ and multi-camera sampling window length $\tau_\text{M}$ to \textbf{the same} as that of temporal sliding windows adopted in SCT and MCT data associations, respectively. In this manner, we can prepare the Siamese metric networks with the exact amount of appearance changes they might encounter in data associations. In practice, we set the data sampling windows lengths as the temporal sliding window lengths, whereas the latter is usually set as the average traveling time inside a single camera (for SCT) or across multiple cameras (for MCT) \cite{ristani2018features,tang2019cityflow}. 
In Section~\ref{sec:results}, we further study the influence of different sampling window lengths.

\subsection{Siamese Metric Network}\label{structure}
For the proposed adaptive affinity module, we replace re-ID feature distances with affinity scores estimated by a Siamese metric network \cite{bromley1993signature}. 
Given the absolute difference between the data pairs, the network learns a binary classifier with 3 hidden layers. 
The metric network outputs a $2$-dim probability distribution $\bm{p}_{ij}=\left(p_{ij}^-,p_{ij}^+\right)$, where $p_{ij}^-$ and $p_{ij}^+$ encode the possibility of the input pair being of different identities or the same identity, respectively. 
The affinity score for the proposed metric is computed by, 
\begin{align}
\label{eq:w_metric}
    a_{ij} = p_{ij}^+-p_{ij}^-. 
\end{align}
This affinity value should be positive if the data pair belongs to the same identity, and negative if otherwise.

During \textbf{training}, the re-ID feature extractor is fixed, and only the metric network is updated with a binary cross-entropy loss. 
During \textbf{testing}, neural network classifiers can easily get over confident~\cite{guo2017calibration} with the prediction, with then turns the affinity to essentially either $-1$ or $1$ in most scenarios. However, such overconfidence can cause trouble for the graph optimization algorithm, as all positives and negatives are treated equally. For this reason, we exert a temperature scaling factor of $0.1$ onto the softmax layer, so as to prevent overconfident outputs. 

\subsection{Discussion}\label{sec:discuss}

\begin{figure}[t]
\centering
\includegraphics[width=\linewidth]{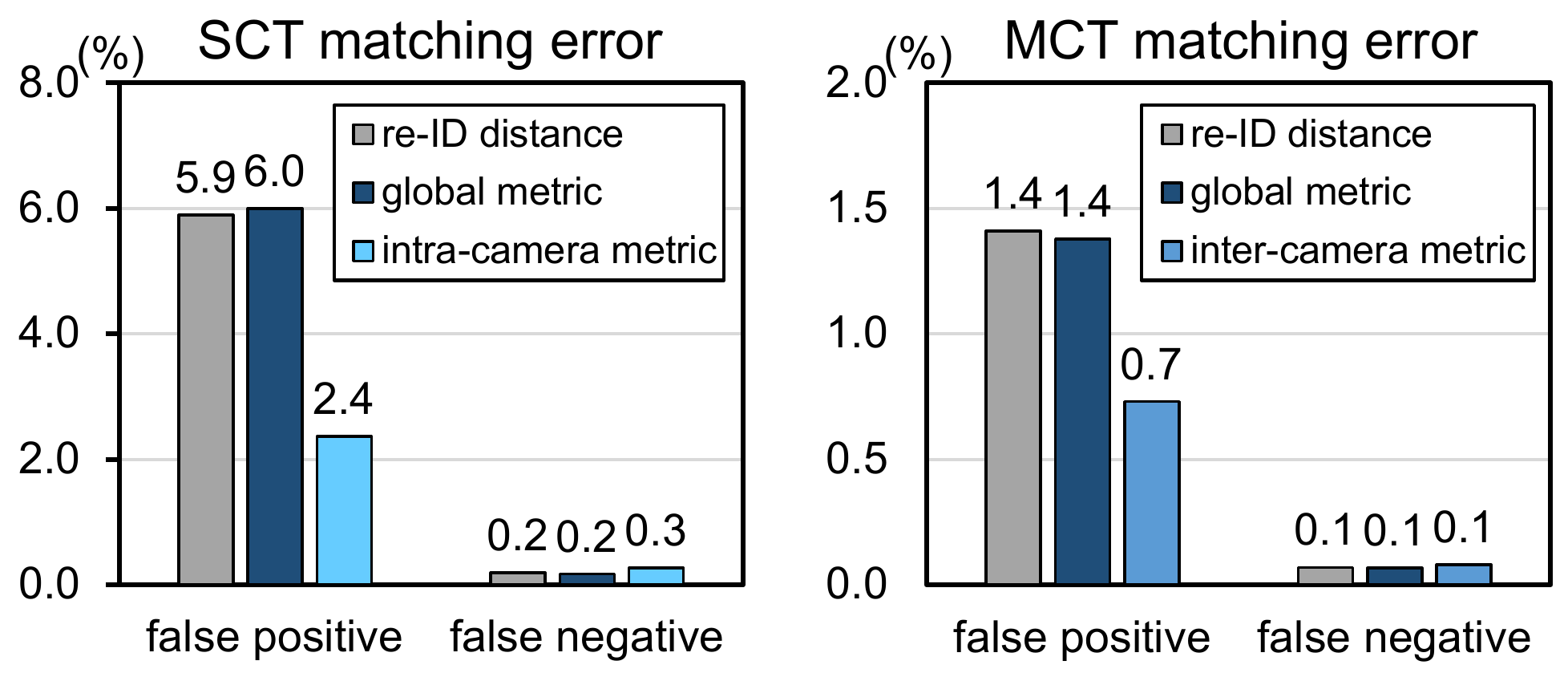}
\caption{
Matching error comparison between the global re-ID metric, ``global metric'' that learns the Siamese metric with global sampling, and the proposed adaptive appearance metric (intra-camera metric and inter-camera metric). We report false positives and false negatives on DukeMTMC validation set.
}
\label{fig:fpfn}
\end{figure}

\textbf{Preliminary experiments on the effectiveness of adaptive affinity.}
We show that the proposed adaptive appearance module can better address the mismatch between affinity and association. In Fig.~\ref{fig:fpfn} (which follows the same experiment design as Fig.~\ref{fig:affinity_dist} and Table~\ref{tab:affinity_dist}. See Section~\ref{secsec:mismatch} for more details), we show the matching errors during SCT and MCT under a similar setting as Fig.~\ref{fig:affinity_dist} and Table~\ref{tab:affinity_dist}. 
Using the proposed inter and intra camera metrics, when compared to the global re-ID distances, we can reduce the false positives significantly while maintaining a similar level of false negatives.

\textbf{Main contributors to performance increases.}
In addition, we show that the improvements come from the focus on the problem size (using temporal sampling window) rather than applying an additional metric network for similarity estimation. To this end, we create ``global metric'', one that adopts the same Siamese network and training mechanism as the proposed adaptive affinity metrics, but without the temporal sampling window. For the global metric, we adopt global data sampling, \ie, setting the temporal window length to infinity and allowing data from all cameras for both positive pairs and negative pairs.
Between ``global metric'' and ``re-ID distance'', we witness no significant differences in either false positives or false negatives, indicating the Siamese metric network architecture makes little difference. 
As such, the drop in false positives should come from the usage of temporal data sampling, which to a certain extent validates that similarity metrics should follow guidance from matching scopes.

\textbf{Extreme cases.} First, under extremely low frame rates, unless they are returning, each target will only appear in one camera once. In this case, SCT is no longer needed and intra-camera metric will not be useful. 
However, since the trajectory continuity still holds, the locality in MCT associations will not be influenced. Thus, the inter-camera metric is still useful. 
Second, the scenario can be in open topology, \ie, targets travel to all cameras at the same probability. This time, the inter-camera metric will fall back to the global metric. However, the SCT associations are still local, and thus the intra-camera metric remains useful.

\section{Experiment} \label{sec:results}

\subsection{Datasets and Evaluation Protocol}
\textbf{CityFlow}~\cite{tang2019cityflow} is a vehicle tracking dataset over an entire city. Specifically, we use the AI-City-2019 challenge MTMCT (track-1) and evaluate on its online test set. CityFlow has a relatively low frame rate (10fps), severe occlusion, and fast-moving vehicles from 40 cameras, spanning over 2km. 

\textbf{DukeMTMC}~\cite{ristani2016performance} is a pedestrian tracking dataset that includes 1080p 60fps videos from 8 cameras on a school campus. Due to some reasons, the 35-minute test partition is no longer available on the MOTchallenge website \cite{leal2015motchallenge}. We report some results tested on the test partition online and the others on the selected validation partition. 
Specifically, we use the first 40 minutes of the training set to train the re-ID feature extractors and affinity Siamese metrics, and the remaining 10 minutes as the validation set. 




\textbf{Evaluation protocol.} 
For \textbf{MTMCT}, following \cite{ristani2016performance}, we use IDF1, IDP, and IDR as evaluation metrics. 
IDF1 is the ratio of correctly identified detections over the average number of ground-truths and computed detections. IDP (IDR) is the fraction of computed detections (ground truth detections) that are correctly identified. As IDF1 considers both false negatives (considered in IDR) and false positives (considered in IDP), we use it as the main evaluation criterion.  
For \textbf{MOT}, we adopt MOTA (multiple object tracking accuracy) as the main criterion following the CLEAR metric \cite{bernardin2008evaluating}. We also report MT (mostly tracked), ML (mostly lost), and IDs (ID switches). 
For \textbf{re-ID}, we adopt the rank-1 accuracy and mean average precision (mAP)~\cite{zheng2015scalable} evaluation protocol.


\begin{table}[t]
\caption{Variants compared in our experiments.}
\label{tab:alias}
\centering
\resizebox{\linewidth}{!}{
\begin{tabular}{l|c|c}
\toprule
Method/variant               & SCT affinity      &  MCT affinity   \\                                                                                                                                                                                                                 \hline
re-ID distance            & re-ID feature distance & re-ID feature distance\\ 
global metric       & global metric     &   global metric \\ 
intra / global       & intra-camera metric & global metric\\ 
global / inter        & global metric  & inter-camera metric  \\ 
adaptive affinity         & intra-camera metric & inter-camera metric   \\ \bottomrule
\end{tabular}
}
\end{table}

\textbf{Variants and notations.} As shown in Table~\ref{tab:alias}, 
``re-ID distance'' uses the Euclidean distance between feature pairs as in Eq.~\ref{eq:w_a} for affinity. 
The other variants adopt the Siamese metric score Eq.~\ref{eq:w_metric} for similarity estimation.  
Similar to experiments in Fig.~\ref{fig:fpfn}, ``global metric'' uses the same Siamese metric network but samples training data pairs randomly from all cameras and all times. 
``adaptive affinity'' is the proposed full system.

\subsection{Implementation Details}\label{implement}

\textbf{Object detector} performance greatly influences the overall tracking-by-detection system~\cite{wang2020towards}. As such, to minimize the influence of different detectors and examine the potential performance increase from our adaptive affinity module, we directly adopt the provided detection results by the datasets: SSD \cite{liu2016ssd} for CityFlow, and OpenPose \cite{cao2018openpose} for DukeMTMC. 

\textbf{Re-ID features.} 
On DukeMTMC, we inspect three globally learned re-ID features, namely, the ID-Discriminative Embedding (IDE)~\cite{zheng2015scalable}, the triplet feature~\cite{hermans2017defense}, and the Part-based Convolutional Baseline (PCB)~\cite{sun2018PCB}, all of which use ResNet-50 \cite{he2016deep} pre-trained on ImageNet~\cite{deng2009imagenet} as the backbone. 
In the following experiments on DukeMTMC, we use the IDE feature in our tracker unless otherwise specified. 


On CityFlow, we use a DenseNet-121~\cite{huang2017densely} based re-ID feature with both softmax and triplet loss. 

\textbf{MTMCT tracker.}
We adopt DeepCC \cite{ristani2018features} as our MTMCT tracker, which adopts hierarchical and iterative associations. 
For hierarchical association, detection bounding boxes are first grouped into tracklets, then into single camera trajectories in SCT. Lastly, in MCT, single camera trajectories are linked across cameras. For iterative association, temporal sliding windows (either within camera or across cameras) are adopted. 
On DukeMTMC, each tracklet has $40$ frames. The temporal sliding window lengths for SCT and MCT are $600$ frames and $2,400$ frames, respectively. 
On CityFlow, we set the tracklet length to $10$ frames. Temporal sliding windows for SCT and MCT are set to $150$ frames and $500$ frames, respectively. 
$\mu_p$ and $\mu_n$ are calculated from the training set in both datasets.

\textbf{Siamese metric learning.} 
We train the Siamese metric network with a learning rate of $1\times10^{-3}$ for the 40 epochs. We adopt the Cosine learning rate scheduler for its fast convergence~\cite{loshchilov2016sgdr}. A cross-entropy loss and a batch size of $64$ are adopted in training. Sampling window lengths are set the same as corresponding matching windows lengths in either SCT or MCT. 
On DukeMTMC, we set $\tau_\text{S}=600$ and $\tau_\text{M}=2,400$. 
On CityFlow, we set $\tau_\text{S}=150$ and $\tau_\text{M}=500$. 

\begin{table}[t]
\caption{
CityFlow online test set results (\%) for multiple camera tracking (AI-City-2019 challenge MTMCT, track-1). Methods with $^*$ include additional data in training, thus yielding higher performance. Method with~$^\dag$ escapes from the tracking-by-detection paradigm and also adopts video object tracking as reference. The proposed adaptive affinity method yields substantial accuracy increase over global re-ID distance and global metric as affinity estimations.
}
\label{tab:cityflow}
\centering
\small
\resizebox{\linewidth}{!}{
\begin{tabular}{l|l|ccc}
\toprule
\multirow{2}{*}{Method} & \multirow{2}{*}{Detector} & \multicolumn{3}{c}{CityFlow test set results} \\ \cline{3-5} 
                        &                           & IDF1            & IDP             & IDR            \\ \hline
team 52                 & -                         & 28.5            & -               & -              \\ 
team 104                & -                         & 33.7            & -               & -              \\ 
team 107                & -                         & 45.0            & -               & -              \\ 
team 36                 & -                         & 49.2            & -               & -              \\ 
team 59$^*$~\cite{tan2019multi}             & Cascade R-CNN~\cite{cai2018cascade}     & 59.9            & -               & -              \\ 
team 97$^*$~\cite{hou2019locality}             & SSD~\cite{liu2016ssd}                       & 65.2            & -               & -              \\ 
team 53        & -          & 66.4            & -               & -              \\ 
team 12$^*$~\cite{he2019multi}             & FPN~\cite{lin2017feature}                       & 66.5            & -               & -              \\ 
team 49$^*$~\cite{li2019spatio}             & FPN                       & 68.7            & -               & -              \\ 
\textbf{team 21}$^{\dag}$~\cite{hsu2019multi}          & Mask R-CNN~\cite{he2017mask}                 & \textbf{70.6}            & -               & -              \\ \hline
re-ID distance          & \multirow{5}{*}{SSD}                       & 56.6            & 53.3            & 60.7           \\ 
global metric           &                        & 57.1            & 54.4            & 60.7           \\ 
intra / global          &                        & 61.2            & 59.1            & 63.9           \\ 
global / inter          &                        & 58.5            & 55.6            & 62.2           \\ 
\textbf{adaptive affinity}       &                        & 63.0          &  60.7            & 66.0           \\  
\bottomrule
\end{tabular}
}
\end{table}

\subsection{Evaluation of the Tracker}
As shown in Table~\ref{tab:cityflow} and Table~\ref{tab:duke_test}, using global re-ID feature distances as affinities, we achieve competitive performance on both CityFlow and DukeMTMC dataset. 
On CityFlow, using the provided SSD~\cite{liu2016ssd} detection results and only the provided data for training, ``re-ID distance'' achieves 56.6\% MCT IDF1, lagging behind the top-performing teams in the AI-City-2019 challenge that either adopt private detector or use additional data in re-ID feature learning. 
On DukeMTMC, using the provided OpenPose~\cite{cao2018openpose} detection results \cite{ristani2018features}, ``re-ID distance'' achieves 91.3\% and 87.4\% for SCT and MCT IDF1 on test (easy); and 83.7\% and 75.4\% for SCT and MCT IDF1 on test (hard), outperforming the previous methods. 

\subsection{Evaluation of the Adaptive Affinity Module} 

\textbf{Improvements over the global re-ID distance.}
We first compare the proposed adaptive affinity module against traditional global re-ID feature distance. Results on CityFlow and DukeMTMC are shown in Table~\ref{tab:cityflow}, Table~\ref{tab:duke_test}, and Table~\ref{tab:duke_val}. 
Going from ``re-ID distance'' to ``adaptive affinity'', we witness consistent and non-trivial improvements on the two datasets. On CityFlow, the proposed adaptive affinity improves MCT IDF1 by +6.4\%. On DukeMTMC test (hard), our method excels the re-ID distance baseline by +2.1\% for SCT and +6.9\% for MCT in terms of IDF1. On the DukeMTMC validation set, ``adaptive affinity'' improves the MCT IDF1 results by +2.4\%, +3.6\%, and +2.3\% using IDE, triplet and PCB features, respectively. 
Such improvements are coherent with our preliminary experiments on matching errors in Fig.~\ref{fig:fpfn}, where the intra-camera metric for SCT association and inter-camera metric for MCT association exhibit lower false positive ratios while maintaining similar false negative ratios. 
Overall, the MTMCT experiments and results demonstrate the effectiveness of the proposed adaptive affinity.

\begin{table}[]
\caption{Preliminary results for MOT scenarios.}
\label{tab:mot}
\centering
\setlength{\tabcolsep}{8pt} 
\begin{tabular}{l|c|c|c|c|c}
\toprule
                  & MOTA   & IDF1   & MT  & ML  & IDs \\ \hline
Re-ID distance    & 64.9\% & 65.7\% & 265 & 124 & 263 \\ \hline
Global affinity   & 64.3\% & 65.2\% & 262 & 126 & 266 \\ \hline
Adaptive affinity & 64.8\% & 66.3\% & 259 & 125 & 264 \\ \bottomrule
\end{tabular}
\end{table}

\begin{table*}[t]
\caption{DukeMTMC online test set results (\%). On both test sets, ``adaptive affinity'' achieves competitive performance.}
  \label{tab:duke_test}
\centering
\resizebox{0.9\linewidth}{!}{
\small
\begin{tabular}{l|l|ccc|ccc|ccc|ccc}
\toprule
\multirow{3}{*}{Method} & \multirow{3}{*}{Detector} & \multicolumn{6}{c|}{DukeMTMC test (easy)}                    & \multicolumn{6}{c}{DukeMTMC test (hard)}                    \\ \cline{3-14} 
                         &                           & \multicolumn{3}{c|}{SCT} & \multicolumn{3}{c|}{MCT} & \multicolumn{3}{c|}{SCT} & \multicolumn{3}{c}{MCT} \\ \cline{3-14} 
                         &                           & IDF1   & IDP    & IDR    & IDF1   & IDP    & IDR    & IDF1   & IDP    & IDR    & IDF1   & IDP    & IDR    \\ \hline
BIPCC~\cite{ristani2016performance}                    & DPM~\cite{felzenszwalb2010object}                       & 70.1   & 83.6   & 60.4   & 56.2   & 67.0   & 48.4   & 64.5   & 81.2   & 53.5   & 47.3   & 59.6   & 39.2   \\ 
MTMC\_CDSC~\cite{tesfaye2019multi}               & DPM                       & 77.0   & 87.6   & 68.6   & 60.0   & 68.3   & 53.5   & 65.5   & 81.4   & 54.7   & 50.9   & 63.2   & 42.6   \\ 
MYTRACKER~\cite{yoon2018multiple}                & DPM                       & 80.3   & 87.3   & 74.4   & 65.4   & 71.1   & 60.6   & 63.5   & 73.9   & 55.6   & 50.1   & 58.3   & 43.9   \\ 
MTMC\_ReIDp~\cite{zhang2017multi}              & DPM                       & 79.2   & 89.9   & 70.7   & 74.4   & 84.4   & 66.4   & 71.6   & 85.3   & 61.7   & 65.6   & 78.1   & 56.5   \\ 
TAREIDMTMC~\cite{jiang2018online}               & Mask R-CNN~\cite{he2017mask}                 & 83.8   & 87.6   & 80.4   & 68.8   & 71.8   & 66.0   & 77.9   & 86.6   & 70.7   & 61.2   & 68.0   & 55.5   \\ 
DeepCC~\cite{ristani2018features}                   & OpenPose~\cite{cao2018openpose}                  & 89.2   & 91.7   & 86.7   & 82.0   & 84.4   & 79.8   & 79.0   & 87.4   & 72.0   & 68.5   & 75.9   & 62.4   \\ 
MTMC\_ReID~\cite{zhang2017multi}               & Faster R-CNN~\cite{ren2015faster}              & 89.8   & 92.0   & 87.7   & 83.2   & 85.2   & 81.2   & 81.2   & 89.4   & 74.5   & 74.0   & 81.4   & 67.8   \\ 
\textbf{StateAware}~\cite{li2019state}               & Faster R-CNN              & 91.8   & \textbf{93.3}   & 90.3   & 86.8   & 88.2   & 85.4   & \textbf{85.8}   & \textbf{93.6}   & 79.2   & 81.3   & \textbf{88.7}   & 75.1   \\ \hline
re-ID distance           & \multirow{3}{*}{OpenPose}                  & 91.3   & 91.8   & 90.9   & 87.4   & 87.8   & 87.0   & 83.7   & 88.8   & 79.1   & 75.4   & 80.0   & 71.3   \\ 
global metric            &                   & 91.3   & 92.2   & 90.4   & 87.7   & 88.6   & 86.8   & 82.7   & 89.2   & 77.1   & 76.2   & 82.2   & 71.0   \\ 
\textbf{adaptive affinity}        &                   & \textbf{92.5}   & 93.0   & \textbf{92.0}   & \textbf{88.6}   & \textbf{89.0}   & \textbf{88.1}   & \textbf{85.8}   & 91.1   & \textbf{81.1}   & \textbf{82.3}   & 87.4   & \textbf{77.8}   \\ \bottomrule
\end{tabular}
}
\end{table*}

\begin{table}[t]
\caption{IDF1 (\%) on the DukeMTMC validation set. Adaptive affinity provides consistent and significant performance increase over multiple re-ID features. }
\label{tab:duke_val}
\centering
\begin{tabular}{l|cc|cc|cc}
\toprule
\multirow{3}{*}{Method} & \multicolumn{6}{c}{DukeMTMC validation set IDF1 results}                                                                                                                         \\ \cline{2-7} 
                         & \multicolumn{2}{c|}{IDE~\cite{zheng2015scalable}}                         & \multicolumn{2}{c|}{triplet~\cite{hermans2017defense}}                         & \multicolumn{2}{c}{PCB~\cite{sun2018PCB}}                        \\ \cline{2-7} 
                         & SCT & MCT & SCT & MCT & SCT & MCT \\ \hline
re-ID distance                 & 86.4                    & 81.4                     & 86.2                    & 80.9                     & 85.8                    & 80.6                    \\ 
global metric                & 85.9                    & 81.6                     & 84.1                    & 79.7                     & 85.4                    & 80.7                    \\ 
intra / global                    & 87.8                    & 83.1                     & 87.6                    & 83.9                     &  87.1                   & 82.4                    \\ 
global / inter                    & 85.9                  & 82.5                 & 84.1                   & 81.4                     & 85.4                    & 82.5                    \\ 
\textbf{adaptive affinity}              & \textbf{87.9}           & \textbf{83.8}            & \textbf{87.9}           & \textbf{84.5}            & \textbf{87.7}           & \textbf{82.9}           \\ \bottomrule
\end{tabular}
\end{table}

In the MOT scenarios, however, we find the adaptive affinity of limited use in our preliminary experiments. We use MOT17 \cite{leal2015motchallenge} as the training set and MOT15 \cite{leal2015motchallenge} as the testing set (we use the training partition that is publicly available) and JDE \cite{wang2020towards} as the tracker. As shown in Table~\ref{tab:mot}, we find the adaptive affinity does not bring significant improvements over the global re-ID distance baseline. In terms of MOTA, adaptive affinity brings a -0.1\%  overall decrease; in terms of IDF1, adaptive affinity brings a +0.6\% overall increase. Coherent with our analysis in Section~\ref{sec:discuss}, targets in MOT scenarios only appear within a single camera and a short time period, which makes the re-ID features learned in MOT scenarios already suitable to the problem scale in MOT data associations. In comparison, for MTMCT scenarios, adaptive affinity brings performance boosts because it can bridge the mismatch global re-ID features (all cameras at all time) and MTMCT data associations (single camera or multiple cameras but within a short time period). Since the mismatch is less pronounced in MOT scenarios, the proposed adaptive affinity is less effective. 

For multiview scenarios (multiple cameras with overlapping fields-of-view focused on the same scenario), when jointly considering multiview information, the problem scales of re-ID feature learning and data association are similar. On Campus and Shelf datasets \cite{belagiannis20143d}, since the targets have large appearance disparities and jointly considering multiple cameras well addresses the occlusion issue, existing methods achieve very high tracking results (\eg, MOTA and IDF1 both around 98\% from \cite{dong2019fast}). Due to the same reasons, we do not report quantitative results on these two datasets, since we believe adaptive affinity would not bring significant improvements. Moreover, like single-view MOT scenarios, adaptive affinity is also less effective in this scenario, as the problem scales are similar. 


\textbf{Comparison with the state-of-the-art methods.}
The proposed ``adaptive affinity'' further improves performance for our tracker and achieves competitive performance on CityFlow while reaching new state-of-the-art on DukeMTMC. 
On CityFlow, our tracker with adaptive affinity module reaches top-6 using the provided SSD detector.
Note that we cannot achieve as competitive results as some of the challenge participants. This is because the test participants focus on building a competitive tracker and adopt various techniques, \eg, using more training data, modeling of vehicle motion and road network in that city, cropping road-side vehicles with prior knowledge, and including single-object tracking results for cross-reference. On the other hand, this work focuses on improving the affinity metric to fit the data association in MTMCT, and is parallel to the mentioned works on building a stronger tracker. 
On DukeMTMC, using the provided OpenPose~\cite{cao2018openpose} detector, on test (easy), we obtain 92.5\% and 88.6\% IDF1 on SCT and MCT, respectively. These numbers are +0.7\% and +1.8\% higher than previous state-of-the-art \cite{li2019state}. On DukeMTMC test (hard), our IDF1 scores are 85.8\% and 82.3\% on SCT and MCT, which translate into a tie and a +1.0\% improvement, respectively. 


\textbf{Main contributors to performance increases.}
There are two possible sources of the improvements in adaptive affinity: the Siamese metric network, and temporal data sampling. Siamese metric network learns affinity using neural network layers, which possibly has an edge over the Euclidean distance for affinity. Temporal data sampling, on the other hand, focuses on the mismatch between affinity and the association problem size, where we believe exist a mismatch. 
To verify the source of improvements and our mismatch arguments, we examine the tracking performances from the ``global metric'' variant, which samples training data globally from all cameras at all times. 
On CityFlow and DukeMTMC, global metric cannot bring constant and significant performance increases. 
For example, global metric brings a mere +0.4\% MCT IDF1 improvement on CityFlow dataset, but fails to improve SCT IDF1 on DukeMTMC test sets. On DukeMTMC validation, global metric sometimes gives performance increases for SCT or MCT, but also sometimes leads to performance drops. 
Overall, the performance differences between global Siamese metric and global re-ID distance are not constant and significant. which agrees with the finding in our preliminary experiments in Fig.~\ref{fig:fpfn}. 
This leaves the temporal data sampling the source of performance increase, which is also supported by the tracking performance increase. On CityFlow, adaptive affinity increases performance by +5.9\% over global metric. On DukeMTMC validation, adaptive affinity also brings consistent performance improvements over global metric, \eg, +2.0\% SCT IDF1 and +2.2\% MCT IDF1 using IDE features. 
In summary, these results verify the temporal data sampling as the source of improvements, further validating the mismatch between global affinities and the local matching nature in SCT and MCT.

\subsection{Variants and Ablation Study}
\label{secsec:variants}

\textbf{Necessity of intra and inter camera metric.} 
In Table~\ref{tab:cityflow} and Table~\ref{tab:duke_val}, we replace the intra and inter camera metric with the global metric, and find both are necessary. When replacing \textbf{intra-camera} metric with the global metric, IDF1 drops by -4.5\%, -2.0\%, and -1.3\% on CityFlow, DukeMTMC SCT, and DukeMTMC MCT (both with IDE features), respectively. As SCT and MCT are conducted in an orderly fashion, changing intra-camera metric for SCT can also lead to performance differences for MCT. 
A similar but smaller accuracy drop can be observed when the \textbf{inter-camera} metric is replaced with the global metric. Specifically, this leads to a -1.8\% IDF1 drop on CityFlow, and -0.1\% and -0.7\% for SCT and MCT on DukeMTMC (using IDE as re-ID features). 
Actually, when changing the inter-camera metric for MCT, the slight performance drops in SCT are due to that some targets returning to the same camera after a long-time departure (much longer than considered problem size in SCT), and is only considered in MCT. 
The removal of the intra-camera metric causes a larger accuracy drop. This is because the problem size differences between MCT and re-ID is smaller compared to that of SCT and re-ID. 
Overall, these results show that both the intra-camera and inter-camera metrics are necessary components in our system. 

\begin{figure}[t]
\centering
\includegraphics[width=\linewidth]{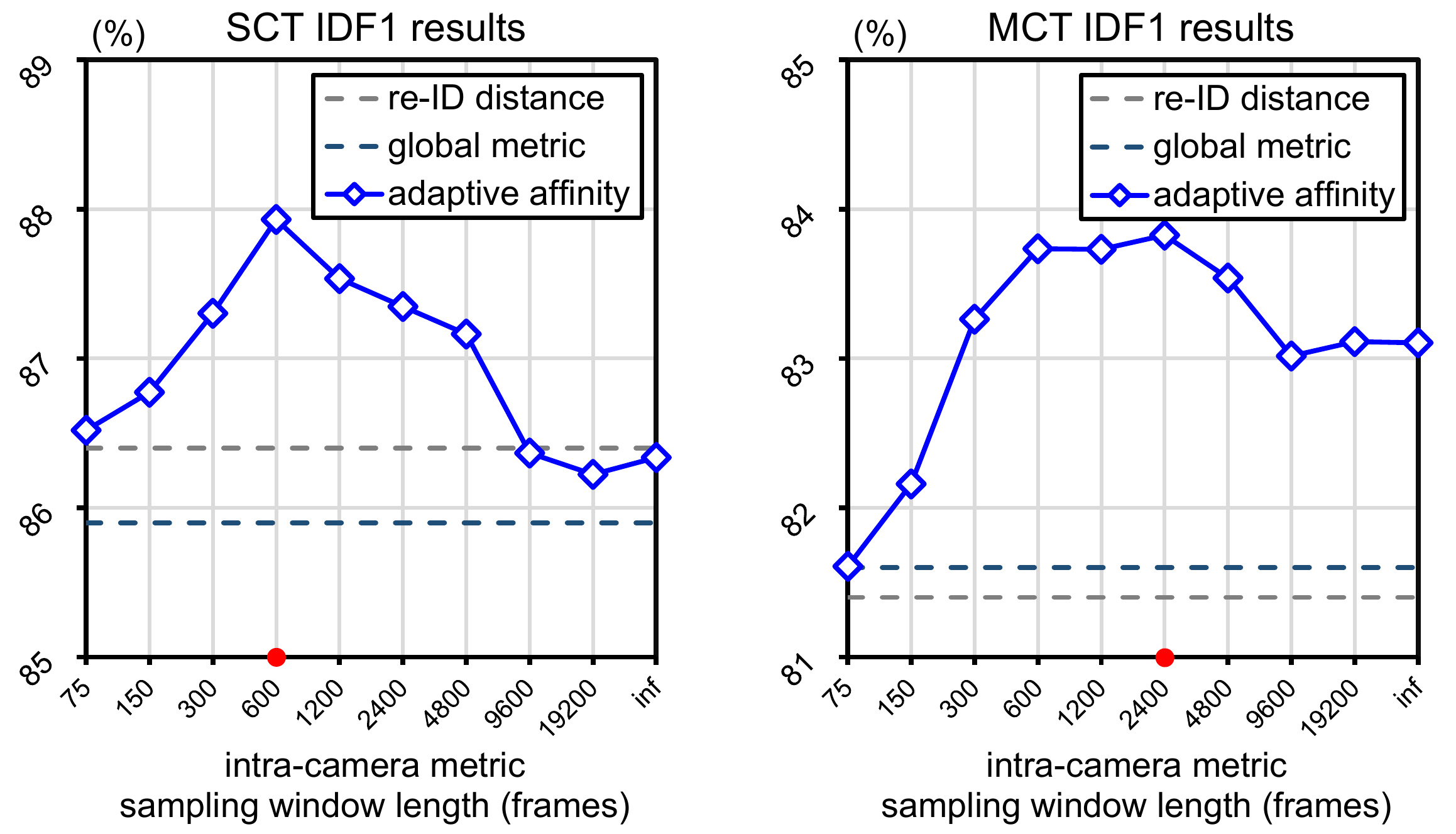}
\caption{Influence of data sampling window lengths for intra and inter camera metrics on DukeMTMC validation set (using IDE features). Red dots highlight the corresponding temporal sliding window lengths for data associations in SCT and MCT, respectively. When data sampling window lengths are set to \textbf{the same} as temporal sliding window lengths, a best match between affinity estimation and data association is achieved, providing highest results. 
}
\label{fig:sampling_window}
\end{figure}

\textbf{Temporal sampling window lengths.} 
In Fig.~\ref{fig:sampling_window}, we evaluate the impact of different data sampling window lengths. 
Too short or too long a temporal window will affect SCT and MCT results. 
Too short a sampling window may significantly reduce the choices of training pairs, and the Siamese metric network cannot learn to deal with the possible appearance changes in association. 
On the other hand, too long a sampling window can include too many possible training pairs, and a large part of which might not really appear in the association problem. 
Adaptive affinity achieves best results in both SCT and MCT when sampling windows are set to the same size as the temporal sliding windows in data associations in SCT and MCT ($600$ and $2,400$), respectively. In this manner, the adaptive affinity modules learn to focus exactly on the possible appearance changes that might appear in associations.

\textbf{Impact of different re-ID features.}
Tracking accuracy with different re-ID features is summarized in Table \ref{tab:duke_val}. Under both the SCT and MCT task, we find that the tracking performance of IDE, triplet, and PCB features similar. This finding is consistent with a previous report \cite{ristani2018features}: improvement in re-ID accuracy can have a diminishing improvement on the MTMCT system. 
The main reason is that the re-ID and MTMCT are two different problems in terms of matching scopes. Re-ID (global matching) deals with all possible appearance variations at once, requiring the features to be overall robust. MTMCT (local matching) deals with limited appearance changes in both SCT and MCT, and the more robust re-ID feature (higher re-ID performance) might not necessarily translate into being more discriminative in the matching problems in MTMCT. 
For example, in MCT, the matching scope within a temporal sliding window might have dozens of images, while that in re-ID has over 10k images. Within a much smaller matching scope, there is less requirement on feature's discriminative ability, and PCB would have a similar matching accuracy with IDE. 
Moreover, MTMCT also has several other components besides feature-based matching. Imperfectness in other components reduces the improvement brought about by the re-ID features. 


\textbf{Computation complexity.} 
The metric network takes 20 minutes to train using one GTX 1080ti GPU. 
During testing, CNN features are extracted with GPU, and affinities the associations are computed on a 3.2Ghz Intel Xeon CPU. 
Overall, on DukeMTMC, the total tracker run time is 1,464 seconds using re-ID feature distances, and 1,537 seconds using the adaptive affinity (an acceptable 5\% increase).

\section{Conclusion}

This paper points out a previously overlooked problem in MTMCT: global re-ID feature distances might not be the best affinity estimation for local matching in either SCT or MCT. We design experiments to verify such misfit, and propose a simple-yet-effective adaptive affinity module for different data associations in MTMCT. Specifically, rather than trying to solve all possible appearance changes, we tailor the affinity metric to focus only on ones that might emerge in data associations in SCT or MCT. With temporal windows originally used for data associations, we introduce a new data sampling method for affinity metric learning. The proposed adaptive affinity introduces significant performance improvements on multiple datasets. In future works, we would like to investigate automatic tuning of sampling window lengths for further improvements.


%



\ifCLASSOPTIONcaptionsoff
  \newpage
\fi



%
\bibliographystyle{IEEEtran}
\bibliography{egbib}

%

\vspace{-10mm}
\begin{IEEEbiography}[{\includegraphics[width=1in,height=1.25in,clip,keepaspectratio]{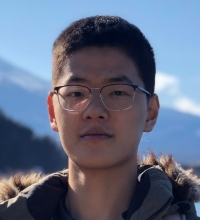}}]{Yunzhong Hou}
received his bachelor degree in electronic engineering from Tsinghua University in 2018. He is now working towards a PhD degree at Australian National University under the supervision of Dr. Liang Zheng and Prof. Stephen Gould. His research interests lies in computer vision and deep learning. 
\end{IEEEbiography}
\vspace{-10mm}

\begin{IEEEbiography}[{\includegraphics[width=1in,height=1.25in,clip,keepaspectratio]{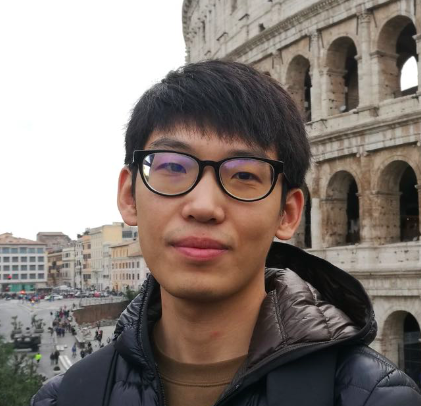}}]{Zhongdao Wang}
received his B.S degree in the Department of Physics at Tsinghua University in 2017. He is now working towards the Ph.D. degree in the Department of Electronic Engineering at Tsinghua University. His research interests include computer vision, pattern recognition and particularly person/face recognition and retrieval.
\end{IEEEbiography}
\vspace{-10mm}

\begin{IEEEbiography}[{\includegraphics[width=1in,height=1.25in,clip,keepaspectratio]{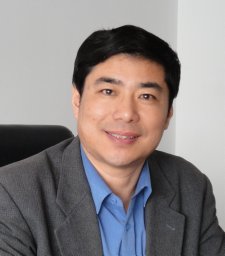}}]{Shengjin Wang}
received the B.E. degree from Tsinghua University, China, in 1985 and the Ph.D. degree from the Tokyo Institute of Technology, Tokyo, Japan, in 1997. From 1997 to 2003, he was a member of Research Staff in the Internet System Research Laboratories, NEC Corporation, Japan. Since 2003, he has been a Professor with the Department of Electronic Engineering, Tsinghua University. He has published over 80 papers on image processing, computer vision, and pattern recognition. His current research interests include image processing, computer vision, video surveillance, and pattern recognition. He is a member of the IEEE and the IEICE.
\end{IEEEbiography}
\vspace{-10mm}

\begin{IEEEbiography}[{\includegraphics[width=1in,height=1.25in,clip,keepaspectratio]{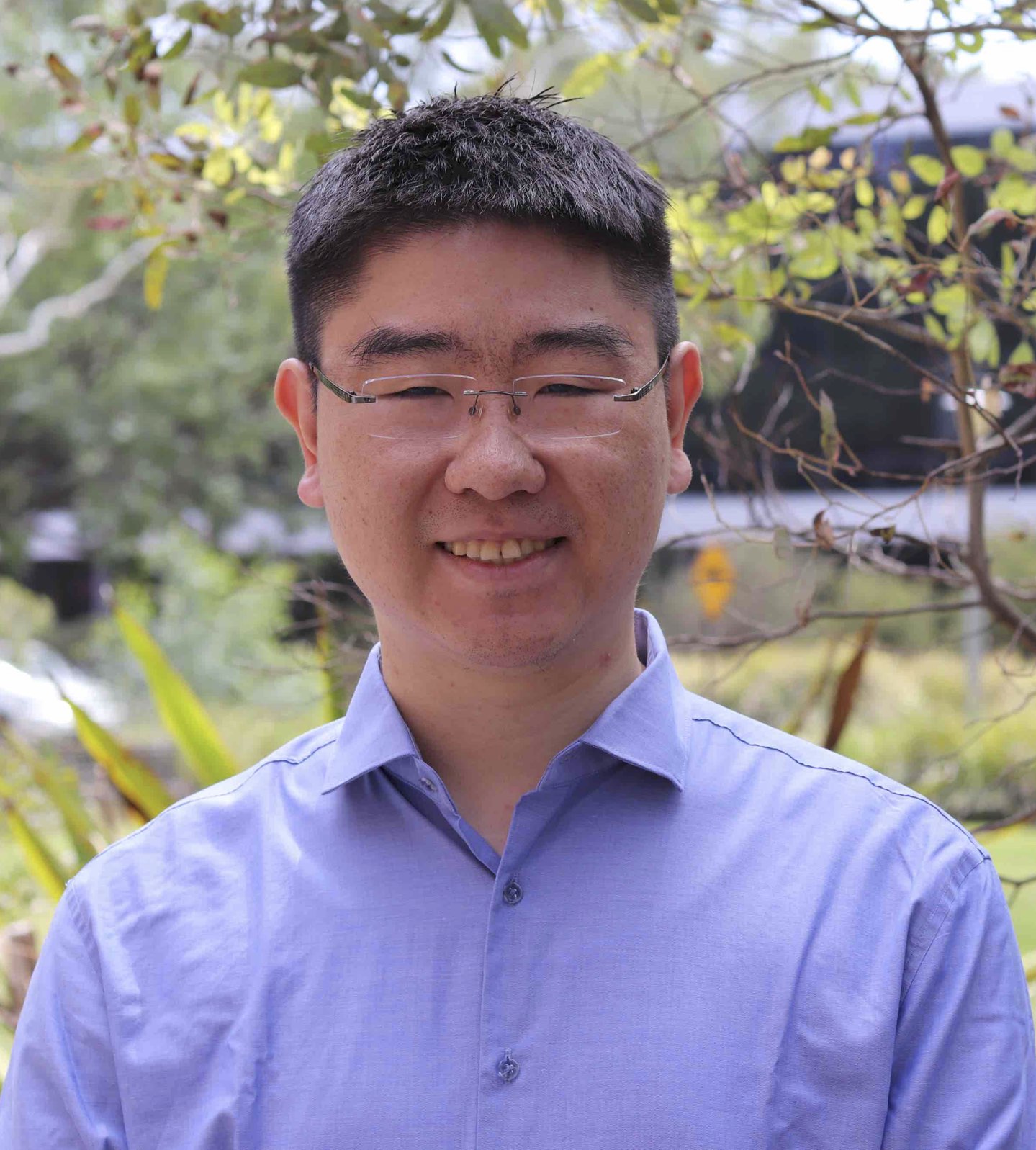}}]{Liang Zheng}
is a Lecturer and a Computer Science Futures Fellow in the Research School of Computer Science, Australian National University. He received the PhD degree in Electronic Engineering from Tsinghua University, China, in 2015, and the B.E. degree in Life Science from Tsinghua University, China, in 2010. He was a postdoc researcher in the Center for Artificial Intelligence, University of Technology Sydney, Australia. His research interests include image retrieval, classification, and person re-identification.
\end{IEEEbiography}

\vfill




\end{document}